%% file: main.tex
\documentclass[10pt,twocolumn,letterpaper]{article}

\usepackage{cvpr}              % To produce the CAMERA-READY version
\usepackage{capt-of}
\usepackage{xspace}
\usepackage{xcolor}
\usepackage{enumitem}
\usepackage{amsmath,amsthm,amssymb,amsfonts,dsfont,pifont,bm,bbm,mathrsfs,mathtools,nicefrac,extarrows,relsize}
\usepackage{algorithm,algpseudocode,listings}
\usepackage{booktabs,multirow,adjustbox,diagbox,threeparttable,tabularray,setspace}
\definecolor{cvprblue}{rgb}{0.21,0.49,0.74}
\usepackage[pagebackref,breaklinks,colorlinks,citecolor=cvprblue,bookmarks=false]{hyperref}
\usepackage{wrapfig}
\usepackage[capitalize]{cleveref}  % Should be loaded after 'hyperref', and works perfectly with 'subfigure'.
\newcommand{\methodname}{AvatarPointillist}
\newcommand{\method}{\texttt{\methodname}\xspace}

\crefname{section}{Sec.}{Secs.}
\Crefname{section}{Section}{Sections}
\crefname{appendix}{App.}{Apps.}
\Crefname{appendix}{Appendix}{Appendices}
\crefname{table}{Tab.}{Tabs.}
\Crefname{table}{Table}{Tables}
\crefname{figure}{Fig.}{Figs.}
\Crefname{figure}{Figure}{Figures}
\crefname{equation}{Eq.}{Eqs.}
\Crefname{equation}{Equation}{Equations}
\crefname{theorem}{Thm.}{Thms.}
\Crefname{theorem}{Theorem}{Theorems}
\crefname{lemma}{Lem.}{Lems.}
\Crefname{lemma}{Lemma}{Lemmas}
\crefname{remark}{Rem.}{Rems.}
\Crefname{remark}{Remark}{Remarks}
\crefname{corollary}{Cor.}{Cors.}
\Crefname{corollary}{Corollary}{Corollaries}
\crefname{algorithm}{Alg.}{Algs.}
\Crefname{algorithm}{Algorithm}{Algorithms}
\definecolor{cvprblue}{rgb}{0.21,0.49,0.74}
\definecolor{lightred}{RGB}{200, 50, 50}   % 浅红色（最佳）
\definecolor{lightblue}{RGB}{50, 100, 200} % 浅蓝色（次佳）
\definecolor{cellred}{RGB}{213, 123, 101}
\definecolor{cellgreen}{RGB}{0, 205, 0}
\definecolor{cellblue}{RGB}{54, 125, 189}
\definecolor{codegreen}{rgb}{0,0.6,0}
\definecolor{codegray}{rgb}{0.5,0.5,0.5}
\definecolor{codepurple}{rgb}{0.58,0,0.82}
\definecolor{backcolour}{rgb}{1.0,1.0,1.0}
\def\paperID{ 2416} % *** Enter the Paper ID here
\def\confName{CVPR}
\def\confYear{2026}

\title{AvatarPointillist: AutoRegressive 4D Gaussian Avatarization}

\author{
Hongyu Liu$^{1,2,*}$ \qquad
Xuan Wang$^{2, \S}$ \qquad
Zijian Wu$^{2}$ \qquad
Yating Wang$^{2}$ \qquad
Ziyu Wan$^{3}$ \\[3pt]
Yue Ma$^{1}$ \qquad
Runtao Liu$^{1}$ \qquad
Boyao Zhou$^{2}$ \qquad
Yujun Shen$^{2}$ \qquad 
Qifeng Chen$^{1,\S}$ \\[8pt]
$^{1}$HKUST \quad 
$^{2}$Ant Group \quad
$^{3}$City University of Hong Kong \\[5pt]
\url{https://kumapowerliu.github.io/AvatarPointillist}
}
\newcommand\nonumfootnote[1]{%
\begingroup%
    \renewcommand\thefootnote{}\footnote{\hspace{-4pt}#1}%
    \addtocounter{footnote}{-1}%
\endgroup%
}
\begin{document}

\twocolumn[{
\renewcommand\twocolumn[1][]{#1}
\maketitle
\begin{center}
    \vspace{-5pt}
    \includegraphics[width=1.0\linewidth]{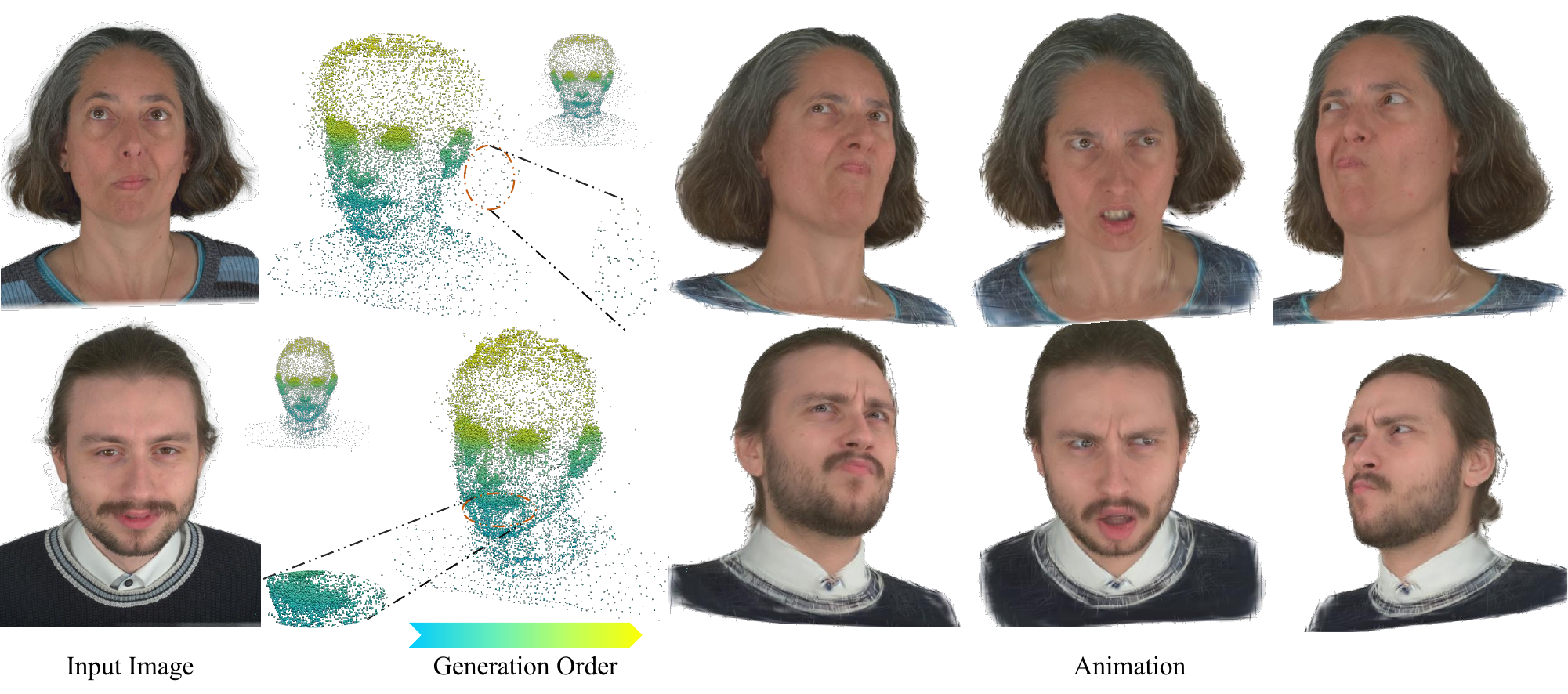}
    \vspace{-15pt}
    \captionsetup{type=figure}
    \caption{\textbf{Gallery of the proposed AvatarPointillist}.The leftmost column shows the input image, the middle column displays the Gaussian point cloud generated by our AR model, and the rightmost column presents the final drivable 4D Gaussian avatar. The generation order proceeds from bottom to top and left to right. It can be seen that our AR model directly models the Gaussian point cloud, allowing it to simulate the adaptive point adjustment capability of Gaussian Splatting to produce precise geometry (e.g., hair and dense beards).
    }
    \label{fig:teaser}
    \vspace{15pt}
\end{center}
}]

\maketitle
\input{sec/0_abstract}

\input{sec/1_intro}

\input{sec/2_related}

\input{sec/3_preliminaries}

\input{sec/4_method}
\input{sec/5_experiments}

\input{sec/6_conclusion}
{
    \small
    \bibliographystyle{ieeenat_fullname}
    \bibliography{main}
}

% WARNING: do not forget to delete the supplementary pages from your submission 
% \input{sec/X_suppl}
   7\end{document}

%% file: sec/0_abstract.tex
\begin{abstract}
We introduce \method, a novel framework for generating dynamic 4D Gaussian avatars from a single portrait image. At the core of our method is a decoder-only Transformer that autoregressively generates a point cloud for 3D Gaussian Splatting. This sequential approach allows for precise, adaptive construction, dynamically adjusting point density and the total number of points based on the subject's complexity. During point generation, the AR model also jointly predicts per-point binding information, enabling realistic animation. After generation, a dedicated Gaussian decoder converts the points into complete, renderable Gaussian attributes. We demonstrate that conditioning the decoder on the latent features from the AR generator enables effective interaction between stages and markedly improves fidelity. Extensive experiments validate that AvatarPointillist produces high-quality, photorealistic, and controllable avatars. We believe this autoregressive formulation represents a new paradigm for avatar generation, and we will release our code  inspire future research. 
\end{abstract}

%% file: sec/1_intro.tex
 \begin{figure}[t!]
    \begin{center}
        \includegraphics[width=1.\linewidth]{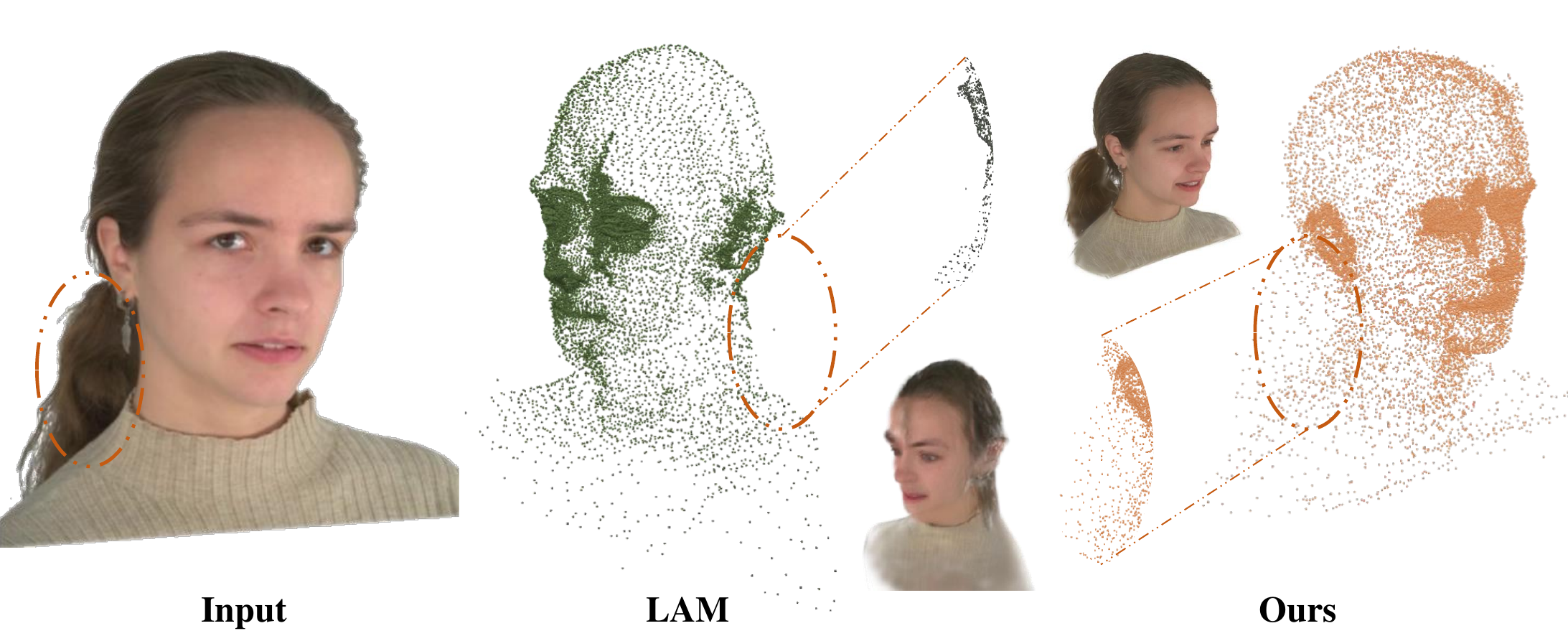}
    \end{center}
    \vspace{-0.7em}
    \caption{ Comparison of different Gaussian point cloud modeling approaches. LAM~\cite{he2025lam} constructs Gaussian point clouds based on a point cloud template, which fails to reconstruct fine details from the image, such as ponytails. In contrast, our method utilizes an AR model to directly model the Gaussian point cloud. It effectively learns the capability to adaptively adjust point density and count, enabling precise modeling. Moreover, we also include final rendering results for comparison. LAM produces distorted geometry and shows noticeable artifacts. }
    \vspace{-5pt}
    \label{fig:motivation}
   
\end{figure}

\section{Introduction}
\label{sec:intro}

The creation of photorealistic and animatable digital humans, often referred to as avatars, is a significant and highly active research area in computer vision and graphics. This technology holds the key to transformative applications in virtual reality (VR), telepresence, filmmaking, and immersive gaming. Broadly, existing approaches to avatar animation can be categorized into two main paradigms: 2D-based animation and 3D-aware (or 4D) avatarization. 2D methods typically operate in the image domain to generate expressive talking heads, while 4D methods focus on building full 3D geometric representations that ensure consistency across varying poses and viewpoints.

Early 2D methods in this domain primarily leveraged Generative Adversarial Networks (GANs)~\cite{karras2019style,goodfellow2020generative,Karras2021} and adopted a warping-then-generation scheme~\cite{siarohin2019animating,siarohin2019first,zakharov2019few,siarohin2021motion,zhao2022thin,guo2024liveportrait,zhang2023metaportrait} to synthesize  facial expressions and pose changes. With the recent emergence of powerful diffusion models~\cite{rombach2021highresolution, ho2020denoising,guo2023animatediff}, a new wave of 2D methods has demonstrated impressive results in generation quality and generalization capabilities~\cite{wei2024aniportrait,xie2024x,ma2024followyouremoji,xu2024hallo}. However, diffusion-based techniques often require substantial computational resources and suffer from long inference times due to the need for multiple sampling steps. More fundamentally, all 2D-based methods lack a sense of 3D structure. This inherent limitation leads to poor handling of extreme pose variations, noticeable geometric distortions, and an inability to render avatars from arbitrary viewpoints. \nonumfootnote{* This work is done partially when Hongyu is an intern at Ant Group.}%
\nonumfootnote{\S~Joint corresponding authors.}%

4D-based methods generate animatable, multi-view-consistent avatars by leveraging 3D geometry. Many approaches use Neural Radiance Fields (NeRF)~\cite{mildenhall2021nerf} as the 3D representation\cite{ma2023otavatar, deng2024portrait4d, zhao2024invertavatar, li2023one, liu2025avatarartist}, achieving good quality but suffering from slow rendering due to NeRF’s inefficiency. Recently, 3D Gaussian Splatting (3DGS)~\cite{kerbl20233d} has emerged as a faster alternative, enabling real-time performance with photorealistic results.
Some methods, such as $\text{GAGAvatar}$~\cite{chu2024gagavatar} and $\text{LAM}$~\cite{he2025lam}, leverage $\text{3DGS}$ for single-image avatar generation, achieving good overall performance but with limited fidelity in fine-grained and identity-specific details. We argue that this issue arises from a fundamental problem in how these methods model the explicit geometry in the 3DGS representation. GAGAvatar~\cite{chu2024gagavatar}, for instance, attempts to lift input 2D features directly into 3D, bypassing a  complete 3D point cloud for representing the head. This design may limit its ability to handle large-angle views and occluded regions, requiring an auxiliary 2D network for final refinement. LAM\cite{he2025lam} addresses this by placing Gaussians directly in a 3D canonical space. However, it relies on a fixed point cloud template (e.g., FLAME vertices\cite{FLAME:SiggraphAsia2017}) and uses a constant number of Gaussians for all subjects. As shown in Fig.~\ref{fig:motivation}, this rigid setup limits the model’s ability to adaptively adjust the density or position of Gaussians to capture subject-specific features like beards or unique hairstyles. As a result, it loses one of the core advantages of 3DGS: adaptive control over point distribution based on geometry. This observation leads to our central question: Can we design a generative model that learns the 3DGS point cloud distribution directly, without relying on a fixed template? Such a model would be free to decide where to place points and how many to use, fully capturing the flexible and adaptive nature that gives 3DGS its power.

% We argue this quality gap arises from a fundamental conflict: these end-to-end generative models require a fixed set of geometric primitives, which directly opposes the inherent dynamic adaptation of the native $\text{3DGS}$ representation. Specifically, these approaches start from a point cloud template (e.g., $\text{FLAME}$ vertices~\cite{FLAME:SiggraphAsia2017} or a points plane~\cite{chu2024gagavatar}) and enforce a fixed number of Gaussians to represent all subjects. This rigid constraint inherently limits the model's ability to adaptively densify or reposition Gaussians for subject-specific features (e.g., beards and unique hairstyles).  In contrast, the effectiveness of original $\text{3DGS}$ fitting can be largely attributed to its dynamic optimization, which adaptively densifies (splits) Gaussians in high-gradient regions and prunes (merges) them elsewhere, this process is crucial to the final rendering quality of 3DGS.  

% In this paper, we introduce AvatarPointillist, a framework that formulates 3DGS avatar generation as an autoregressive (AR) sequential task. This point-by-point generation approach intuitively facilitates the modeling of complex geometry (i.e., mesh~\cite{hao2024meshtron}) and, more critically, enables the model to dynamically adjust the point distribution. It learns to place Gaussians with greater density and more precise positioning in complex regions, thereby achieving significantly higher fidelity. 

In this paper, we propose AvatarPointillist, a novel framework that directly tackles this challenge by casting 3DGS avatar generation as an autoregressive (AR) sequential task. Unlike existing methods that rely on fixed templates, our approach learns to generate the 3DGS point cloud distribution from scratch. This point-by-point generation paradigm fully embraces the adaptive and dynamic nature of 3DGS, enabling the model to adjust the spatial distribution of Gaussians on the fly—placing points with higher density and finer precision in geometrically complex regions. 
To train this, we first employ a fitting procedure~\cite{qian2024gaussianavatars} to construct Gaussian point data for each identity in a 4D avatar dataset~\cite{kirschstein2023nersemble}, creating dynamically densified 3DGS data with animation binding for each subject. We then quantize this data and train a decoder-only Transformer using a next-token prediction objective. To incorporate identity-specific features, we introduce cross-attention mechanisms that inject identity embeddings into the Transformer. Our model effectively learns to fit this structured data, enabling it to adaptively adjust the spatial distribution and scale of Gaussian points based on the input image, thus supporting high-quality and identity-aware avatar generation.

 Once the sequential generation of the point cloud geometry is complete, we utilize a separate Transformer based Gaussian decoder to translate these points into their full Gaussian parameters (e.g., color, opacity, etc.) for rendering. We found that by conditioning this decoder on the latent features from the AR generator, we significantly enhance the final rendering quality. Comprehensive experiments demonstrate that our method significantly outperforms all baselines, both quantitatively and qualitatively. We believe this exploration of autoregressive generation for explicit avatar geometry represents a promising new direction for the community.

% Unlike methods that generate all points at once (one-shot) from a fixed template, our method sequentially predicts the position of the next 3D Gaussian point based on the point cloud already generated.  The point-by-point generation approach intuitively facilitates the modeling of complex geometry, resembling the densification behavior observed in standard 3D Gaussian Splatting (3DGS) fitting, and the AR manner also demostrate the powerfull .  Specifically, we use the fitting method output  as our dataset, and 

% The point-by-point generation approach intuitively facilitates the modeling of complex geometry, resembling the densification behavior observed in standard 3D Gaussian Splatting (3DGS) fitting.  Specifically, this AR manner enables the model to dynamically adjust the point distribution—placing Gaussians with greater density and more precise positioning in complex regions—thereby achieving significantly higher fidelity. Moreover, our method is not constrained by a fixed total point count due to the autoregressive strategy, allowing it to adapt the geometric complexity to each individual subject. Moreover, as our AR model generates each point, it simultaneously predicts its binding information. This binding associates each 3DGS point with a corresponding face on the template mesh, similar to ~\cite{Qian_2024_CVPR}. This linkage ensures that when the mesh deforms, the 3DGS points move accordingly, resulting in a fully animatable 4D head avatar.

%% file: sec/2_related.tex
\section{Related Works}
\label{sec:related}
This section briefly reviews related work, including 2D and 3D-aware avatar generation methods, as well as recent approaches on autoregressive geometry generation.

\subsection{2D-Based Animatable Avatar}

Image-driven talking head synthesis has seen rapid advancements in recent years, particularly within the 2D generation paradigm~\cite{siarohin2019animating,siarohin2019first,zakharov2019few,burkov2020neural,drobyshev2022megaportraits,siarohin2021motion,gong2023toontalker,yin2022styleheat,wang2023progressive,xu2024vasa,zhang2023metaportrait,ma2024followyourpose,liu2023human}. Many work leverages Generative Adversarial Networks (GANs) to produce realistic speaking face videos, 
% typically following a pipeline that first encodes identity from a reference image, 
then applies motion-driven warping, and finally renders the output frames. To guide the warping process, different motion cues such as facial landmarks~\cite{zakharov2019few,siarohin2019first}, depth maps~\cite{hong2022depth}, and latent representations~\cite{burkov2020neural} have been utilized to ensure accurate expression and motion transfer from the driving source. With the emergence of diffusion-based generative models, several recent approaches~\cite{xie2024x, ma2024followyouremoji, wei2024aniportrait} have incorporated pre-trained diffusion backbones into the one-shot talking face generation pipeline. These methods benefit from strong priors learned on large-scale image datasets.
% , which help them generalize well to diverse reference portraits, including stylized or out-of-domain faces. 
However, due to their inherently two-dimensional modeling assumptions, these approaches often struggle with large pose variations, leading to visible geometric artifacts. Furthermore, they lack explicit 3D awareness, making view control and consistent head movement synthesis particularly challenging.

\subsection{3D-Aware Animatable Avatar }

\paragraph{Fitting-Based Methods.}
Given a monocular video as input, some per-subject optimization method utilizing representations like meshes~\cite{grassal2022nha}, NeRFs~\cite{gafni2021nerface, zielonka2023insta, xu2023avatarmav,Wang_2025_CVPR, 10.1145/3641519.3657512, liu2025headartist}, SDFs~\cite{zheng2022imavatar}, points~\cite{zheng2023pointavatar}, and 3D Gaussians~\cite{xiang2024flashavatar, chen2024monogaussianavatar}. However,   the optimization-based nature of these methods often leads to overfitting on the input viewpoint, resulting in poor extrapolation to novel views. Some research~\cite{hong2022headnerf, mildenhall2021nerf, yu2024one2avatar} leverages large-scale multi-view datasets~\cite{kirschstein2023nersemble,   pan2024renderme360, yang2020facescape, buehler2024cafca, xu2025gphm, zheng2024headgap} to learn rich, generalizable priors for geometry and appearance. 
% This has enabled high-fidelity models based on NeRF~\cite{hong2022headnerf, mildenhall2021nerf, yu2024one2avatar} and 3D Gaussian representations~\cite{xu2025gphm, zheng2024headgap} that achieve state-of-the-art realism. 
However, these approaches are fundamentally fitting-based—they are primarily designed to reconstruct or adapt a model to a specific subject, often from scratch. While they can achieve impressive reconstruction quality, their procedures are typically rigid and lack flexibility for broader use cases. More recently, methods such as CAP4D~\cite{taubner2024cap4d} and GAF~\cite{tang2024gaf} have introduced diffusion models to synthesize multi-view images from a single input portrait, which are then used to drive the avatar fitting process. Although this strategy improves identity generalization, it still requires considerable time for optimization, limiting its practicality in real-time or one-shot scenarios.

\paragraph{End-to-End Methods.}
To address the need for generalization, end-to-end methods learn a powerful prior from large-scale monocular~\cite{xie2022vfhq} or multi-view datasets~\cite{kirschstein2023nersemble, martinez2024codec}, enabling them to generate an animatable avatar from a single or very few images.   A significant milestone in this direction is the advent of Neural Radiance Fields (NeRF)~\cite{mildenhall2021nerf, chan2022efficient, yu2023nofa, li2023one, li2024generalizable, ma2023otavatar, zhuang2022mofanerf, trevithick2023real, chu2024gpavatar, ye2024real3d}, which support high-fidelity 3D reconstruction and fine-grained camera control. Some approaches incorporate 3D supervision from monocular 3D face reconstruction~\cite{danvevcek2022emoca,deng2019accurate,feng2021learning} or synthetic multi-view data\cite{deng2024portrait4d,deng2024portrait4dv2, liu2025avatarartist} for better perfomance.   NeRF-based pipelines have been widely integrated into one-shot talking head generation frameworks, improving the realism and 3D alignment of the synthesized results.  More recently, GAGAvatar~\cite{chu2024gagavatar}, LAM~\cite{he2025lam}, and Avat3R~\cite{kirschstein2025avat3r} demonstrated the effectiveness of 3D Gaussian Splatting (3DGS)~\cite{kerbl3Dgaussians} in this context, offering faster rendering while preserving high visual quality. However, these methods still suffer from some limitations. GAGAvatar~\cite{chu2024gagavatar}, for instance, requires an auxiliary neural network for refinement and its 2D-to-3D lifting strategy struggles to realistically model unseen regions. While LAM~\cite{he2025lam} addresses these particular issues, it is constrained by a fixed-template point cloud~\cite{FLAME:SiggraphAsia2017}. This static topology inherently limits its fidelity, as it cannot adaptively adjust the Gaussian count to match subject-specific features. Avat3R~\cite{kirschstein2025avat3r}, on the other hand, is not a one-shot method, requiring multiple input images, and its network must be re-executed to generate the Gaussian splatting for each new expression. In contrast, our method, AvatarPointillist, addresses these limitations by formulating the task as an autoregressive (AR) generative process. As a one-shot generative model, it is not constrained by a fixed template or topology. This AR approach allows our model to dynamically and adaptively adjust the Gaussian distribution and total count, enabling the high-fidelity synthesis of complex, subject-specific features.

 \begin{figure*}[t!]
    \begin{center}
        \includegraphics[width=0.95\textwidth]{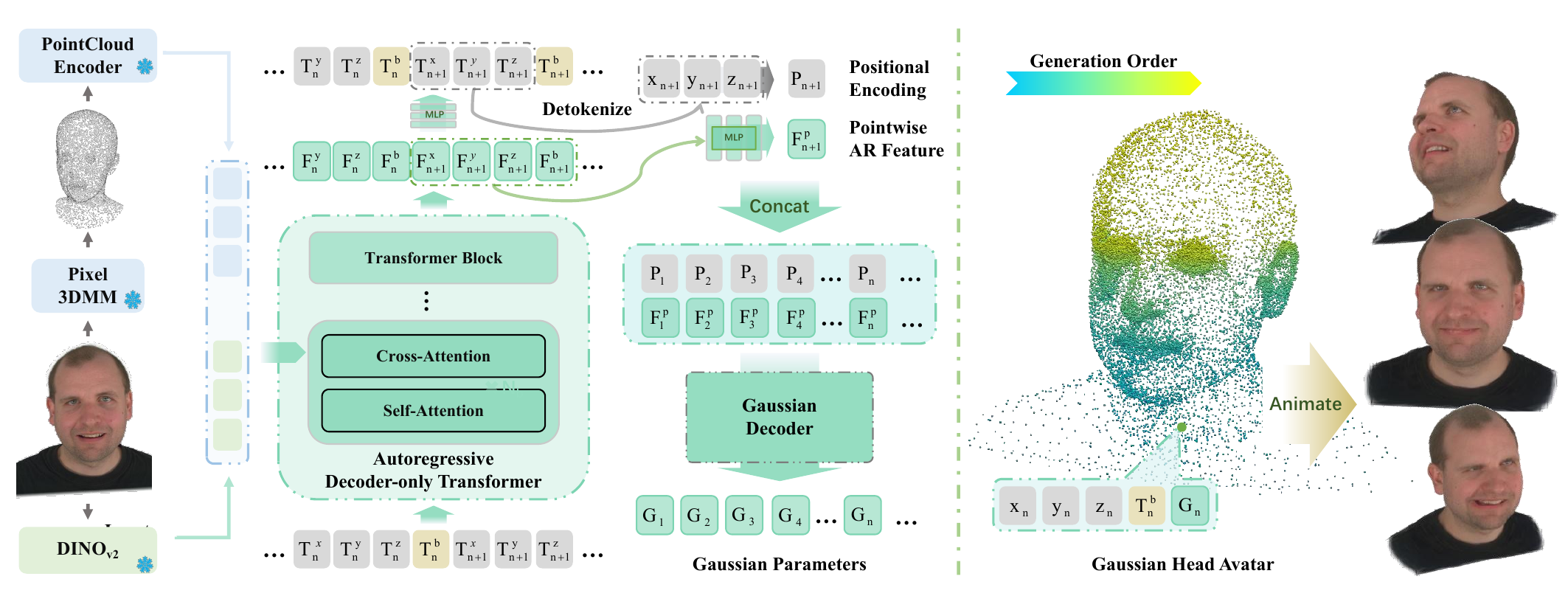}
    \end{center}
    \vspace{-0.7em}
\caption{ Overview of our framework. It consists of two modules: an autoregressive (AR) model for Gaussian geometry generation and a Gaussian Decoder for predicting rendering attributes. The AR model takes image features from DINOv2~\cite{oquab2023dinov2} and point cloud features as input. The point cloud feature extract via Pixel3DMM~\cite{giebenhain2025pixel3dmm} and a point cloud encoder~\cite{zhao2023michelangelo}. The AR model is trained to generate a Gaussian point cloud via next-token prediction, where each point is represented by four quantized tokens $(T_n^x, T_n^y, T_n^z, T_n^b)$ corresponding to coordinates and binding information. After generation, the tokens are de-quantized to obtain the actual coordinates. We then combine the positional embeddings $P_n$  with the internal features  $F_n^p$  from the AR model as input to the Gaussian Decoder to predict the final accurate Gaussian attributes. Finally, the result is animated using Linear Blend Skinning (LBS) and the binding information.
 }
    \label{fig:pipeline}
    \vspace{-0.9em}
\end{figure*}

\subsection{Autoregressive Geometry Generation}
Inspired by the profound success of autoregressive (AR) models in natural language processing~\cite{vaswani2017attention,brown2020language}, a significant trend has emerged in applying sequential generation techniques to 3D geometry. This paradigm typically treats a 3D shape (e.g., a mesh or point cloud) as a sequence of discrete tokens. A common strategy involves a two-stage process: first, a Vector Quantized Autoencoder (VQ-VAE)~\cite{van2017neural} learns a discrete vocabulary of geometric features; second, a Transformer~\cite{vaswani2017attention} is trained to autoregressively predict the next token in the sequence. This approach is famously demonstrated by MeshGPT~\cite{siddiqui2024meshgpt} for triangle meshes, and similar concepts have been applied to point clouds~\cite{sun2020pointgrow} and implicit representations~\cite{mittal2021autosdf}. To overcome the challenge of modeling long sequences, more recent methods like Meshtron~\cite{hao2024meshtron} and ARMesh~\cite{lei2025armesh} propose hierarchical or coarse-to-fine AR generation, significantly improving the fidelity of the resulting geometry.

%% file: sec/3_preliminaries.tex
\section{Preliminary}
\label{sec:Preliminary}
In this section, we provide a brief overview of some essential prerequisites that are closely related to our AvatarPointillist in Section~\ref{sec:method}. We first introduce our 3DGS data structure and explain how we construct the training data. Then,  we present how we quantize the data to make it compatible with AR training.

\subsection{Data Construction}
We build our training data using the   GaussianAvatars~\cite{xiang2024flashavatar} method and the Nersemble dataset~\cite{kirschstein2023nersemble}. Specifically, for each identity in Nersemble, we first fit a complete GaussianAvatars model. This method creates a 3DGS representation where each Gaussian is bound to a specific face of the FLAME mesh~\cite{FLAME:SiggraphAsia2017}. As described in their paper, a 3D Gaussian is  static in the local space of its parent triangle but dynamic in the global space as the triangle moves. For each Gaussian, the model defines its location $\mu$, rotation $r$, and scaling $s$ in this local space. At rendering time, these properties are converted to the global space using the face's transformation (rotation $R$, translation $T$, and scale $k$):
\begin{equation*}
    r' = Rr,\quad \mu' = kR\mu + T,\quad s' = ks.
\end{equation*}
We refer the reader to the original paper~\cite{xiang2024flashavatar} for additional implementation details. To construct our training data, we use the canonical FLAME mesh of each identity to compute the corresponding global canonical Gaussian point cloud. Specifically, let $N$ be the total number of Gaussians in the point cloud, the final point cloud is defined as $P$ as:
\begin{equation}
P = (x_1, y_1, z_1, b_1, x_2, y_2, ...x_N, y_N, z_N, b_N).
\end{equation}
Here, $x_n, y_n, z_n$ are the global coordinates of the Gaussian in the canonical space for each point, and $b_n$ is the binding index, which indicates the FLAME face to which the point is attached.

\subsection{Quantization and Order of Coordinates}
Following the approach introduced in~\cite{hao2024meshtron}, we adopt a specific ordering strategy for our Gaussian point cloud.  We establish this order by sorting all points in a given cloud based on their coordinate values. The primary sorting key is the vertical y-axis, followed by the z-axis, and finally the x-axis (a yzx sort order). This fixed sorting strategy ensures that identical point clouds will always produce identical input sequences for our model.  

In addition to the point cloud coordinates, we structure the sequences using three reserved token types  similar to~\cite{hao2024meshtron}: Start-of-Sequence (S), End-of-Sequence (E), and Padding (P). For each sequence, we prepend a block of 4 start-of-sequence (S) tokens and append a block of 4 end-of-sequence (E) tokens. This design choice reflects the fact that each point consists of four values: the 3D coordinates (x, y, z) and a binding index. Grouping the special tokens in blocks of four ensures structural consistency within the sequence representation.

Our autoregressive model requires discrete tokens as input. We therefore convert our continuous point coordinates into a discrete format using quantization. This is achieved by dividing the coordinate space into a fixed number of bins. The number of bins determines the granularity of the resulting geometry, creating a trade-off between precision and computational load. We found that 1024 quantization levels (similar to strategies in prior work~\cite{hao2024meshtron}) provide an effective balance between model accuracy and efficiency for representing our Gaussian points.  Finally, after quantization, our point cloud $P$ is flattened  into a single integer sequence $T$ for the autoregressive model:
\begin{equation}
T = (T_1^x, T_1^y, T_1^z, T_1^b, \dots, T_N^x, T_N^y, T_N^z, T_N^b).
\end{equation}
Here, each coordinate $T_n^x, T_n^y, T_n^z$ is a discrete token in the range $[0, 1023]$, corresponding to our $1024$ quantization levels. The binding token $T_n^b$ is offset to occupy a distinct part of the vocabulary, defined as $T_n^b = b_n + 1024$. Given that $b_n$ is the original face index (with a maximum of $10144$ faces, so $b_n \in [0, 10143]$), the binding tokens $T_n^b$ fall within the range $[1024, 11167]$.

%% file: sec/4_method.tex
\section{Method}
\label{sec:method}
We aim to develop a method that generates a  4D Gaussian animatable avatar from a single source image $I_s$, driven by the motion of a target individual $I_t$. In Section~\ref{sec:ar}, we introduce an autoregressive mechanism for predicting the  canonical 3D Gaussian Splatting  point cloud. Based on the output of this AR model, a  Gaussian decoder is employed to infer the attributes of each point (e.g., position, scale, and rotation). The input to the AR model consists of both the previously generated 3DGS point and the model’s learned implicit features, as described in Section~\ref{sec:gd}. Furthermore, since our AR model also predicts the binding between each point and the template mesh, the generated canonical 3D Gaussian representation can be animated by deforming it with the mesh motion (see Section~\ref{sec:ea}). An overview of the  pipeline is illustrated in Figure~\ref{fig:pipeline}. We now describe each component in detail.

\subsection{Autoregressive Model}
\label{sec:ar}

The core structure of our AR model is a decoder-only Transformer, the  architecture is shown in Figure~\ref{fig:pipeline}. Specifically, our Transformer  contains several layers, where each layer contains a cross-attention layer, a self-attention layer, and a feed-forward network.

For injecting the input image information, we use  DINOv2~\cite{oquab2023dinov2}  to extract the feature of the input image directly. Meanwhile, since our AR model  focuses on point cloud generation, we also use an  off-the-shelf 3D face reconstruction model~\cite{giebenhain2025pixel3dmm} to get the FLAME parameters. We then use these parameters to get the  sample   vertices from the FLAME mesh  and use a point cloud encoder~\cite{zhao2023michelangelo} to obtain their features. Finally, the DINOv2 feature and point cloud feature are concatenated and injected into our decoder via the cross-attention layers.

With our Transformer, the output is generated by sequentially predicting each  token $T_n$ based on its conditional probability given all previously generated tokens $T_{<n}$: $p(T_n | T_{<n})$. The complete point cloud with binding information is represented as a sequence of $4N$ tokens (i.e., $N$ points with 4 quantized tokens each for $x$, $y$, $z$, and binding). The joint probability of the entire sequence $T$ is modeled as:
\begin{equation}
   p(T) = \prod_{n=1}^{4N} p(T_n | T_{<n}).
\end{equation}
The entire training process uses the standard cross-entropy loss for next-token prediction.

\subsection{Gaussian Decoder}
\label{sec:gd}

Once the AR model generates the complete output sequence, we use a Transformer-based Gaussian decoder to predict the full set of Gaussian parameters (as shown in Fig.~\ref{fig:pipeline}). First, we detokenize the token sequence to recover the original coordinates $(x, y, z)$ for each point. Similar to LAM~\cite{he2025lam}, these coordinates are passed through a positional encoding~\cite{mildenhall2021nerf} and an MLP to produce a per-point geometric feature, $P_n$. Importantly, we found that the  inherent hidden states from the AR Transformer are also crucial for improving generation quality (see Sec.~\ref{sec:abl}). We extract the final hidden state sequence $F$ from AR model:
\begin{equation}
    F = (F_1^x, F_1^y, F_1^z, F_1^b, \dots, F_N^x, F_N^y, F_N^z, F_N^b).
\end{equation}
Since four tokens correspond to a single 3D point, an MLP is used to assemble these four corresponding hidden features ($F_n^x, F_n^y, F_n^z, F_n^b$) into a single, comprehensive AR  feature, $F_n^p$.

Finally, the two features for each point, the geometric feature $P_n$ and the AR feature $F_n^p$ are  concatenated and used as the input to the Gaussian Decoder. This decoder then outputs the final attributes for each Gaussian point $k$:
$c_k \in \mathbb{R}^3$, opacity $o_k \in \mathbb{R}$, scale $s_k \in \mathbb{R}^3$, rotation $R_k \in \mathrm{SO}(3)$, and a positional offset $\Delta p_k \in \mathbb{R}^3$. More specifically, this predicted offset $\Delta p_k$ is added to the canonical point positions, allowing the model to make fine-grained geometric adjustments to better capture the target person's geometry.

\subsection{Expression Animation}
\label{sec:ea}

Our AR model predicts accurate binding information for each point, enabling us to animate the canonical 3D point cloud directly using vertex-based Linear Blend Skinning (LBS) and corrective blendshapes, in a manner similar to the FLAME model.  
First, we interpolate vertex-specific attributes from the FLAME mesh to our output points via barycentric interpolation. For each point $\mathbf{p}_i \in \mathbb{R}^3$, we determine its corresponding triangle $f_i$ on the FLAME mesh with the binding information and retrieve the triangle's vertex indices. Using the vertex positions, we compute the barycentric coordinates $(b_0, b_1, b_2)$  of $p_i$ with respect to the triangle. These coordinates are then used to interpolate the FLAME attributes defined at the vertices. The interpolated LBS weights $ \hat{\mathbf{w}}_i$  and expression blendshapes $ \hat{\mathbf{s}}_i  $ for point $p_i$ are computed as:
\begin{equation}
\begin{aligned}
\hat{\mathbf{w}}_i &= b_0 \mathbf{W}_0 + b_1 \mathbf{W}_1 + b_2 \mathbf{W}_2 \\
\hat{\mathbf{S}}_i &= b_0 \mathbf{S}_0 + b_1 \mathbf{S}_1 + b_2 \mathbf{S}_2
\end{aligned}
\label{eq:interpolated_attributes}
\end{equation}
where $ \mathbf{W}_j $ and $ \mathbf{S}_j  $ are the LBS weights and expression directions at vertex $ j \in \{0,1,2\} $ in the corresponding triangle $f_i $ of the FLAME mesh.

Once equipped with these interpolated properties, our Gaussian avatar is now fully rigged and can be driven by the standard FLAME deformation process using pose ($\boldsymbol{\theta}$) and expression ($\boldsymbol{\psi}$) parameters.

\subsection{Loss Functions and Training Strategy}
\label{sec:loss}
Our model is trained in  two stages. We first optimize the  AR model for sequential generation. After this stage is complete, we freeze the AR model and separately train the Gaussian Decoder using a combination of rendering losses.

\subsubsection{ Autoregressive Model Training}
The AR Transformer is trained first, akin to a standard language model. The objective is to accurately predict the next token $T_n$ in the quantized sequence. We optimize this stage using a standard  Cross-Entropy (CE) loss.

 \begin{figure*}[t!]
    \begin{center}
       \includegraphics[width=0.85\textwidth]{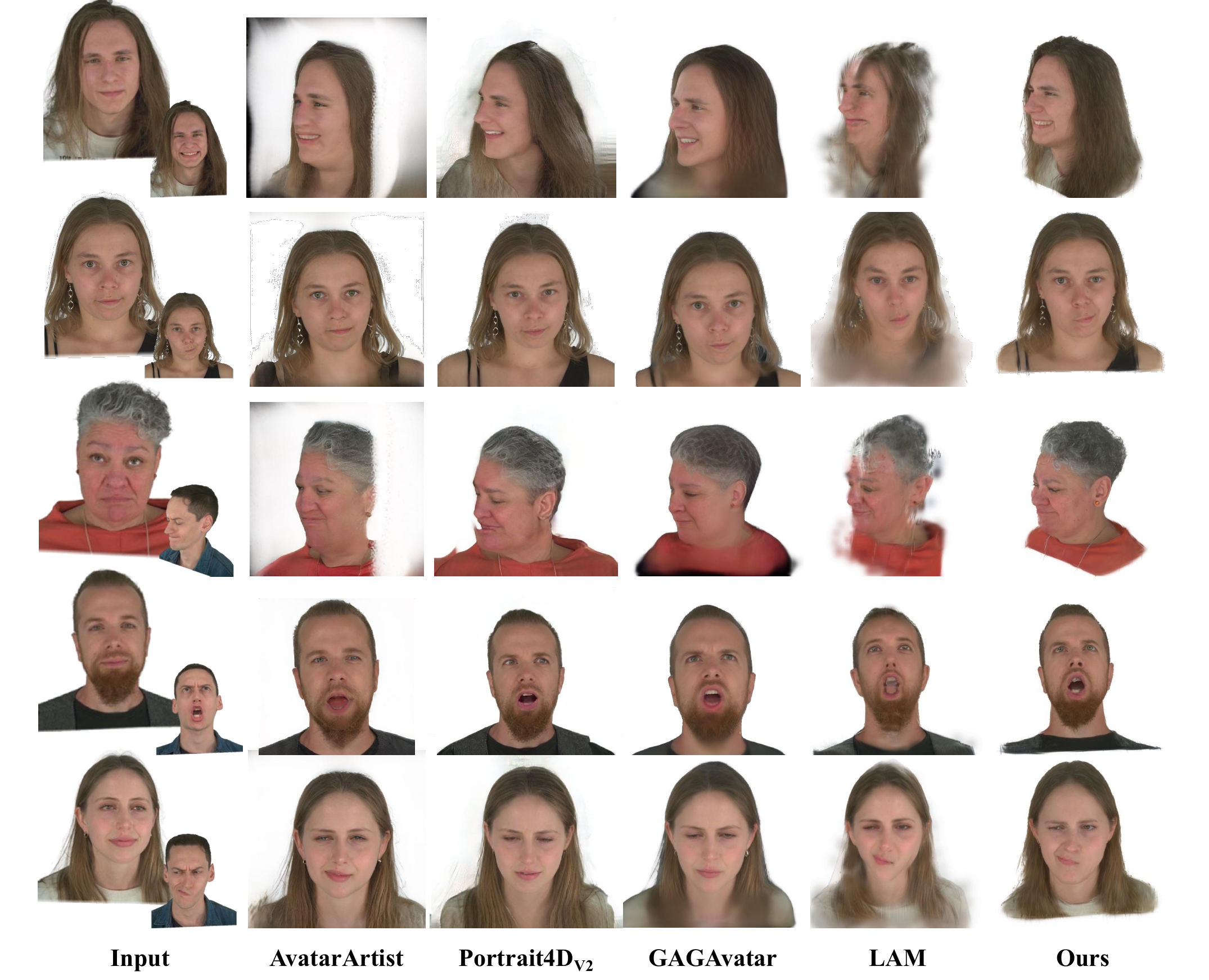}
    \end{center}
    \vspace{-0.7em}
\caption{Qualitative comparison with state-of-the-art  methods. The leftmost column shows the input images, with the target image displayed in the bottom-right corner. The first row presents self-reenactment results, while the remaining three rows show cross-reenactment results. Our method demonstrates superior performance in expression and pose consistency, as well as better identity preservation compared to other approaches.}
    \label{fig:compare}
    \vspace{-0.7em}
\end{figure*}

\subsubsection{Gaussian Decoder Training}

For each identity in the dataset, we use the trained AR model to generate the latent sequence $T$ and hidden states $F_n^p$, which are fed into the Gaussian Decoder to predict 3D Gaussian Splatting (3DGS) attributes. The decoder is optimized by comparing the rendered image $I_r$ with the ground-truth view $I_{gt}$ using a combination of photometric and perceptual losses. Specifically, we employ an $L_1$ loss to ensure pixel-wise color consistency, SSIM to preserve structural similarity, LPIPS to enhance perceptual quality, and a regularization term   applied to constrain the predicted offset. The overall objective is defined as:

\begin{equation}
\begin{aligned}
\mathcal{L}_{\text{total}} =\;& 
\lambda_{L1}\mathcal{L}_{L1} + \lambda_{SSIM}\mathcal{L}_{SSIM} \\
&+ \lambda_{LPIPS}\mathcal{L}_{LPIPS} + \lambda_{Reg}\mathcal{L}_{Reg}
\end{aligned}
\end{equation}

We empirically set the weights to $\lambda_{L1} = 1$, $\lambda_{SSIM} = 0.5$, $\lambda_{LPIPS} = 0.1$ and $\lambda_{Reg} =0.1$.

%% file: sec/5_experiments.tex
\begin{table*}[t]
\centering
\caption{Quantitative evaluation of state-of-the-art methods and our approach on the NeRSemble dataset~\cite{kirschstein2023nersemble}. 
${\downarrow}$ indicates lower is better while ${\uparrow}$ indicates higher is better.
\textbf{\textcolor{lightred}{Red}} highlights the best result, and \textbf{\textcolor{lightblue}{Blue}} highlights the second-best result.}
\resizebox{0.8\linewidth}{!}{
\begin{tabular}{l|llll|llll}
\toprule
\multicolumn{1}{c}{\multirow{2}{*}{Method}} &  \multicolumn{4}{c}{Self reenactment}  & \multicolumn{4}{c}{Cross reenactment} \\ 
\cline{2-9} 
 \multicolumn{1}{c}{} & LPIPS $\downarrow$ &  FID $\downarrow$ & AKD $\downarrow$  & APD $\downarrow$ & FID $\downarrow$ & CLIP $\uparrow$ & AKD $\downarrow$ & APD $\downarrow$\\
 \midrule
 Portrait4Dv2~\cite{deng2024portrait4dv2}       &  0.20                 & 123.02            & 5.32              &34.53             & 191.13                 &0.63                     & 11.94                   &  \textbf{\textcolor{lightred}{142.93}}             \\
 AvatarArtist~\cite{liu2025avatarartist}                     &   0.21                  & 118.94          & 6.87            & 39.58               & \textbf{\textcolor{lightblue}{175.69}}             & 0.61                       & 9.32               &  187.31   \\ 
 LAM~\cite{he2025lam}      &  0.24                  & 136.01             &  4.37               &  61.83               & 238.54                    & 0.54                        &  \textbf{\textcolor{lightblue}{6.68}}                      & 210.23             \\
 GAGAvatar~\cite{chu2024gagavatar}      &  \textbf{\textcolor{lightblue}{0.18}}                  & \textbf{\textcolor{lightblue}{111.76}}                &  \textbf{\textcolor{lightblue}{3.93}}                 &   \textbf{\textcolor{lightblue}{27.94}}                 &181.22                     & \textbf{\textcolor{lightblue}{0.71}}                       & 10.01                       & 170.12               \\
\midrule
Ours                                           & \textbf{\textcolor{lightred}{0.15}}            &\textbf{\textcolor{lightred}{95.18}}     &\textbf{\textcolor{lightred}{2.38}}       & \textbf{\textcolor{lightred}{22.86}}                 &\textbf{\textcolor{lightred}{160.74}}                     &\textbf{\textcolor{lightred}{0.75}}                        &\textbf{\textcolor{lightred}{5.93}}                & \textbf{\textcolor{lightblue}{153.13}}       \\ 
\bottomrule
\end{tabular}} 
\label{tab:experiment}
\vspace{-0.1in}

\end{table*}

\section{Experiments}
% In this section, we first detail our experimental setup, including our implementation details, the baselines used for comparison, and the evaluation metrics. Then, we present the experimental and analytical results, covering both quantitative and qualitative comparisons. Finally, we present our ablation study to analyze our model and validate the effectiveness of our contributions. More results are provided in the supplementary material.

In this section, we first describe our experimental setup, including implementation details, baselines, and evaluation metrics. Then, we present quantitative and qualitative results. Finally, we conduct an ablation study to validate our model and contributions. Additional results are available in the supplementary material.

\begin{figure*}[t!]
    \begin{center}
        \includegraphics[width=0.8\linewidth]{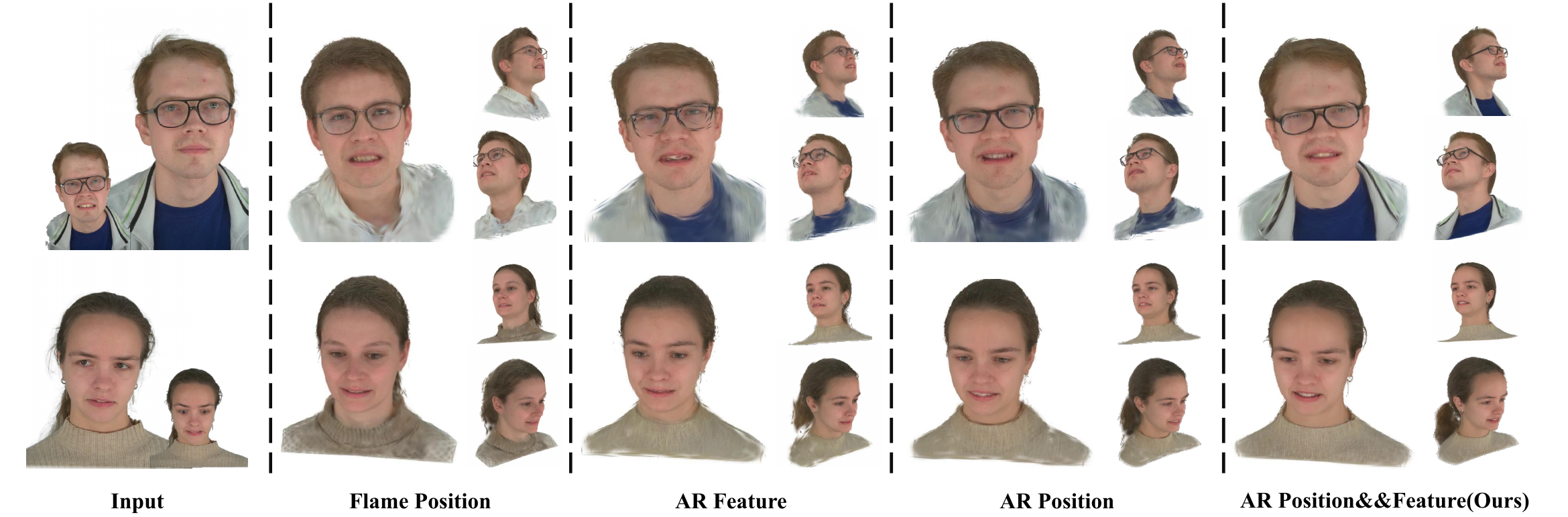}
    \end{center}
    \vspace{-0.7em}
    \caption{\textbf{Visualization of ablation study on input setting of Gaussian decoder.} The leftmost column shows the input. The FLAME Positions baseline, similar to the LAM method, uses the canonical FLAME mesh vertices as a template and only applies decoder-predicted offsets to deform this template into a final Gaussian point cloud. Pointwise AR Feature refers to using only the AR features  ($F_n^p$) without positional information, while Positional Encoding uses only the point embeddings ($P_n$) without AR features. }
    \label{fig:ab_decoder}
    \vspace{-0.8em}
\end{figure*}

\subsection{Experimental Setup}

\paragraph{Implementation Details.}
We train our model on the NeRSemble dataset~\cite{kirschstein2023nersemble}, which features a total of 419 identities. We randomly select 25 of these identities to form our test set, using the remainder for training. To generate the training data for our AR method, we first fit all identities using the GaussianAvatars~\cite{qian2024gaussianavatars} method.
During the training of the autoregressive model, we utilize the AdamW optimizer~\cite{loshchilov2019adamw} with a learning rate of 1e-4. The autoregressive model is trained on 16 NVIDIA H20 GPUs for 50K steps with a batch size of 4. Since the point cloud sequences are very long, we adopt the truncated training strategy from~\cite{hao2024meshtron}to enhance efficiency. Specifically, the input token sequence is first partitioned into fixed-size context windows, with padding applied to any segments of insufficient length. Then, we utilize a sliding window mechanism to shift the window step-by-step and train each windowed segment accordingly. We set the window size to 12000. For the Gaussian Decoder training, we also use the Adam optimizer and train for 12500 steps on 8 NVIDIA H20 GPUs with a batch size of 4. For more details, please refer to the supplementary material.

{\flushleft \bf Baselines.}We compare our method with recent state-of-the-art, single-image-guided 4D avatar reconstruction models, including two NeRF-based methods (AvatarArtist~\cite{liu2025avatarartist} and Portrait4Dv2~\cite{deng2024portrait4dv2}) and two Gaussian Splatting-based methods (LAM~\cite{he2025lam} and GAGAvatar~\cite{chu2024gagavatar}).

{\flushleft \bf Evaluation Metrics.} To evaluate perceptual quality, we adopt LPIPS~\cite{zhang2018unreasonable} and FID~\cite{heusel2017gans}. Expression accuracy is measured using the Average Keypoint Distance (AKD)\cite{lugaresi2019mediapipe}, while pose consistency is assessed by the Average Pose Distance (APD), with pose parameters extracted following\cite{10477888}. For identity preservation, we employ CLIPScore~\cite{radford2021learning} as our ID metric.

\subsection{Qualitative Results} 
As shown in Figure~\ref{fig:compare}, we provide qualitative comparisons for self-reenactment and cross-reenactment tasks. The first column shows the source image and the target pose (inset). We randomly select diverse viewpoints for a thorough evaluation. The top two rows shows self-reenactment results , and the rest show cross-reenactment  . Among baselines, LAM shows clear artifacts, especially in complex facial areas. AvatarArtist works for small pose changes but struggles with larger ones. Portrait4Dv2 and GAGAvatar produce coherent results but often have expression mismatches and over-smoothed hair. In contrast, our method generates more realistic and consistent reenactments, with better alignment in pose and expression. It also preserves fine details like hair texture and facial contours, resulting in sharper and more identity-accurate outputs.

% \subsection{Quantitative Results} The quantitative results are summarized in Table~\ref{tab:experiment}. We evaluate our method on the test set of the NeRSemble dataset~\cite{kirschstein2023nersemble} under both self-reenactment and cross-reenactment settings. For self-reenactment, the source image is randomly selected from intermediate frames of the input sequence, ensuring minimal occlusion. For cross-reenactment, we use a fix motion sequence as the target motion for all methods. All baseline methods are provided with the same source image for each identity to ensure a fair comparison. We also align the input camera views as closely as possible across methods and use each method's own tracking pipeline to obtain ground-truth poses, further ensuring fairness in evaluation.  As shown in Table~\ref{tab:experiment}, our method consistently outperforms existing baselines across all evaluation metrics in both self-reenactment and cross-reenactment tasks, demonstrating its superior capability in capturing identity-preserving expressions and accurate motion transfer.

\subsection{Quantitative Results}
Table~\ref{tab:experiment} summarizes the quantitative results on the test set of the NeRSemble dataset~\cite{kirschstein2023nersemble} under both self- and cross-reenactment settings.
For self-reenactment, the source image is randomly chosen from intermediate frames with minimal occlusion. For cross-reenactment, a fixed motion sequence is used as the target across all methods. All baselines use the same source image per identity, and input camera views are aligned across methods. We also align the input camera views as closely as possible across methods and use each method's own tracking pipeline to obtain ground-truth poses, further ensuring fairness in evaluation.  As shown in Table~\ref{tab:experiment}, our method consistently outperforms all baselines across metrics in both tasks, showing superior identity preservation and motion transfer accuracy.

\begin{table}[t]
\centering
\caption{Ablaiton study on the NeRSemble dataset~\cite{kirschstein2023nersemble}. ${\downarrow}$ indicates lower is better while ${\uparrow}$ indicates higher is better. }
\resizebox{0.85\linewidth}{!}{
\begin{tabular}{l|llll}
\toprule
\multicolumn{1}{c}{Method} & LPIPS $\downarrow$ & FID $\downarrow$ & AKD $\downarrow$ & APD $\downarrow$ \\
 \midrule
Flame Position    &0.23            &   120.34            &   4.82&            41.22             \\
AR Feature     &0.22            &   110.93            &   5.89&            32.96             \\
AR Position    &0.19            &   103.80            &   5.81&            41.49           \\
\midrule
Ours &             \textbf{0.15} &   \textbf{95.18}&   \textbf{2.38}&       \textbf{22.86}            \\ 
\bottomrule
\end{tabular}
}
\label{tab:ablation}
\vspace{-0.2in}

\end{table}

\subsection{Ablation Study}
\label{sec:abl}
% We conduct ablation studies to evaluate the contribution of the autoregressive model in constructing high-quality Gaussian point clouds. In addition, we analyze how different input settings affect the performance of the Gaussian decoder, further demonstrating the effectiveness of the AR model.

{\flushleft \bf Effectiveness of autoregressive Model.} We compare our full method with a baseline called FLAME position, which adopts a static-topology approach similar to LAM~\cite{he2025lam}. This baseline skips our AR model and directly uses 3D vertices from the canonical FLAME mesh as input to the Gaussian decoder, which then predicts offsets to refine these fixed points. As shown in Figure~\ref{fig:ab_decoder}, this static point cloud fails to capture subject-specific geometry and shows limited resemblance to the input image. It struggles with complex regions like hair and cannot adaptively allocate points to important areas. Additionally, since it relies only on point embeddings without rich per-point features, the rendered results often lack identity consistency.
In contrast, our AR model generates the 3D point cloud directly, allowing more accurate geometry reconstruction. The AR features also help the decoder produce higher-quality Gaussian attributes.

{\flushleft \bf Effectiveness of Input Setting of Gaussian Decoder.} 
We further analyze the impact of different inputs to the Gaussian Decoder, as shown in Figure~\ref{fig:ab_decoder}. Using only the final AR hidden state $F_n^p$ (Pointwise AR Feature) yields suboptimal results due to the lack of spatial information. Using only de-quantized 3DGS coordinates $P_n$ (Positional Encoding) performs better by providing spatial context, but misses the semantic richness of the AR features.
Our full method combines both $P_n$ and $F_n^p$, allowing the decoder to leverage spatial guidance and deep semantic cues, resulting in more accurate attribute prediction and the best overall quality.

%  \begin{figure}[t!]
%     \begin{center}
%         \includegraphics[width=0.9\linewidth]{fig/ablation_ar.pdf}
%     \end{center}
%     \vspace{-0.7em}
%     \caption{\textbf{Visualization of ablation study on  AR model.} The leftmost column shows the input. The FLAME Positions baseline, similar to the LAM method, uses the canonical FLAME mesh vertices as a template and only applies decoder-predicted offsets to deform this template into a final Gaussian point cloud. However, this approach fails to capture fine geometric details (e.g., braids) and struggles with identity preservation, as it lacks fine-grained per-point features and relies solely on positional embeddings. }
%     \label{fig:ab_ar}
%     \vspace{-0.6em}
% \end{figure}

%% file: sec/6_conclusion.tex
\section{Conclusion}\label{sec:conclusion}
We propose AvatarPointillist, a novel framework for one-shot 4D Gaussian avatar generation. At its core is an autoregressive model that learns to generate Gaussian point clouds point by point, removing fixed topology constraints. This enables dynamic control over the number and placement of Gaussians, focusing more on complex, identity-specific areas and fully exploiting the adaptive nature of 3D Gaussian Splatting (3DGS).
Our two-stage architecture feeds the AR model’s output and hidden features into a Gaussian decoder to predict high-quality rendering attributes. Experiments show that AvatarPointillist outperforms prior methods in both quantitative metrics and visual quality. We believe this autoregressive approach offers a promising direction for explicit 3D avatar generation.

\section{Acknowledgment}
The work was supported by HKUST under grant number WEB25EG01.
\clearpage